Master's Thesis

# CNN-based Multi-In-Multi-Out Model for Efficient Spatiotemporal Prediction

Department of Artificial Intelligence Convergence

Graduate School, Chonnam National University

Hyeonseok JIN

August 2025

Master's Thesis

# CNN-based Multi-In-Multi-Out Model for Efficient Spatiotemporal Prediction

Department of Artificial Intelligence Convergence
Graduate School, Chonnam National University

Hyeonseok JIN

August 2025

# CNN-based Multi-In-Multi-Out Model for Efficient Spatiotemporal Prediction

Department of Artificial Intelligence Convergence
Graduate School, Chonnam National University

Hyeonseok Jin

Supervised by Professor Kyungbaek KIM

A dissertation submitted in partial fulfillment of the requirements for the **Master of Science**

Committee in Charge :

Manje Kim ____________________

Yeongjun Cho ____________________

Kyungbaek Kim ____________________

August 2025

# Content

# List of Figures

# List of Tables

# CNN-based Multi-In-Multi-Out Model for Efficient Spatiotemporal Prediction

Hyeonseok Jin

Department of Artificial Intelligence Convergence
Graduate School Chonnam National University
(Supervised by Professor Kyungbaek Kim)

(Abstract)

Recently, Convolutional Neural Network (CNN) or Transformer architecture based models have been proposed to overcome the limitations of Recurrent Neural Network (RNN) based models in spatiotemporal prediction. These models prevent the inefficiency of parallelization limitation due to the sequential properties and stacked error due to the recursive method, and show high performance. Novertheless, there are still some challengies. First, CNN based models have difficulty considering global information due to the local properties of the kernel, and their performance is limited. In addition, information is mixed because the time axis is combined with the channel axis of the image for processing. Models based on Transformer architecture have high complexity due to the self-attention calcuation and take a long training time. In this paper, we propose a new structure model called CNN-based Multi-In-Multi-Out model for Efficient Spatiotemporal Prediction (MIMO-ESP) to overcome these limitations. MIMO-ESP considers global information and significantly improves complexity by configuring a Transformer architecture based on CNN. In addition, it treats the time axis as an independent axis without combining it, and effectively considers spatiotemporal information together by applying dilation. This structure makes MIMO-ESP efficient and high performance. Extensive experiment results on three promising benchmark datasets which including video, traffic, and precipitation prediction tasks demonstrate that the usefulness of MIMO-ESP due to the achieved competitive efficiency while outperforming existing models. Furthermore, the ablation study results demonstrate the usefulness of the components of MIMO-ESP, emphasizing the potential of the proposed approaches.

# 1. INTRODUCTION

## A. Backgound

Spatiotemporal prediction is the task of learning meaningful features from past data including temporal and spatial information and accurately predicting future. It extracts complex spatial, temporal, and spatiotemporal correlations in a self-supervised manner using unlabeled data [1], and can play a key role in intelligent systems [2]. Spatiotemporal prediction is receiving much attention with the advancement of deep learning, a promising approach to automatically identifying spatial, temporal, and spatiotemporal correlations [3]. This growing interest has applications in a wide range of fields, including weather prediction [4-6], traffic prediction [7-9], action recognition [10-12], and robotics [13], making numerous contributions. Although deep learning methodologies have shown incredible results, spatiotemporal data used for prediction is mainly in the form of 5-dimensional tensor, including batch, channel, time, height, and width as shown in Fig.1.1 (c). A typical example is video data, which is difficult to analyze and process because of the data size is large and complex [14] than others, including time-series and image, as shonw in Fig.1.1 (a) and (b). Due to the high computational complexity caused by this, efficiency is an important factor as well as predict quality.

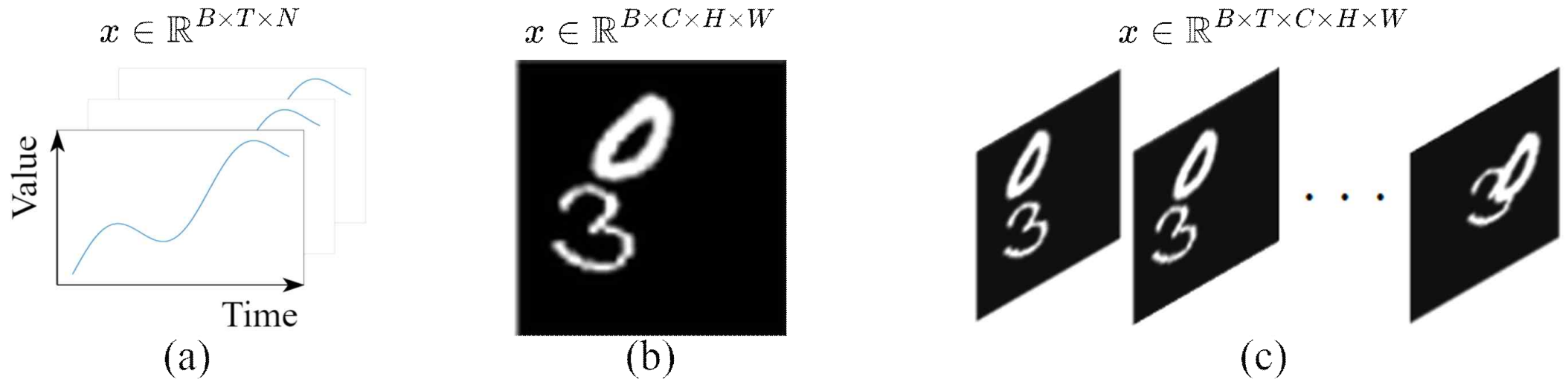

Fig. 1.1. Examples of dimensions per each data structure.

Since the advent of ConvLSTM [15], which extended the concept of Fully-Connected LSTM (FC-LSTM) [16], the Recurrent-based Single-In-Single-Out (SISO) model which combination of Convolutional Neural Network (CNN) that extract spatial feature and Recurrent Neural Network (RNN) that extract temporal feature has become mainstream for spatiotemporal prediction. In recent years, interesting methods based on this have been proposed [12, 17-26]. However, RNNs that extract temporal feature have some limitations. Due to their sequential properties, they have to wait until the prediction for the previous time step is completed in order to perform prediction, which cause limitations is parallelism [27, 28]. In addition, their high computational complexity limits their widespread use in real world applications [29, 30]. And predicting multiple time step in a recursive manner can degrades the performance due to stacked errors [31].

To overcome these limitations, Recurrent-free Multi-In-Multi-Out (MIMO) models that exclude RNNs have been proposed [1, 28, 30-33] and are receiving much attention. MIMO models are based on CNN or Vision Transformer (ViT) [34]. These models receive input data with multiple time steps and predict the future with multiple time steps at once. The efficient structure and excellent parallelism of the encoder-decoder overcome the inherent limitations of RNN. The remarkable efficiency and prediction performance prove the effectiveness of the MIMO methodology. Nevertheless, several challenges still exist and there's a room for improvement.

CNN-based MIMO models have difficulty considering global information. Due to the local properties of the kernel, there are limitations in accessing global information, and generally, multiple CNN layers are stacked to consider global information, but the receptive field increases slowly [35]. In addition, these models integrate the temporal axis and treat them similarly to the channel axes, which mix spatial and temporal features, can resulting in inefficiency and reduced prediction quality [36]. ViT, which emerged due to the success of Transformer [37], is a powerful method in the field of computer vision and is good at considering global information due to its self-attention mechanism. However, in the field of spatiotemporal prediction that deals with high-dimensional data such as 5-dimensional tensors, the computational complexity of the self-attention mechanism is fatal.

### B. Objective

To improve MIMO models and perform efficient spatiotemporal prediction, our goal is propose spatiotemporal prediction model which can consider global information based on CNN. For this, we propose CNN-based Multi-In-Multi-Out model for Efficient Spatiotemporal Prediction (MIMO-ESP). MIMO-ESP adopts a Transformer-style CNN architecture to consider global information and extends it with two attention blocks and one feedforward block to capture spatial and spatiotemporal information. Two attention blocks, Spatial Attention Block (SA-Block) and SpatioTemporal Attention Block (STA-Block) apply attention to spatial and spatiotemporal information, respectively, to enable the model to focus on important information for spatiotemporal prediction. Specifically, SA-Block processes spatiotemporal data similarly to images and applies a CNN with large sized kernels to capture global spatial information. STA-Block efficiently learning spatiotemporal features by applying gradually increasing dilation to the time axis without mixing time and channel. Feedforward block effectively complement the outputs of two attention blocks by doubling and restoring the dimension of hidden layers. Our contributions are summarized as follows:

- We propose a MIMO-ESP which can consider global spatial and

spatiotemporal information. It can efficiently perform spatiotemporal prediction because based on Transformer-style CNN.

- We conduct extensive experiments using three promising spatiotemporal datasets. Experimental results show that MIMO-ESP achieves competitive performance and can be applied to various spatiotemporal prediction fields. In addition, the ablation study results emphasize the usefulness of the components of MIMO-ESP.

# 2. RELATED WORKS

## A. Recurrent-based SISO models

Recurrent-based SISO models use a structure that combines CNN and RNNs as shown in Fig. 2.1. Data corresponding to each time step is input into recurrent cell and processed sequentially. ⊗ and ⊕ denotes hadamard product and element-wise add, respectively. multiple Each recurrent cell contain convolution operation * to extract spatial features from each frame, and performs spatiotemporal prediction by learning and updating cell state $C$ and hidden state $H$ according to the sequence order in RNNs manner.

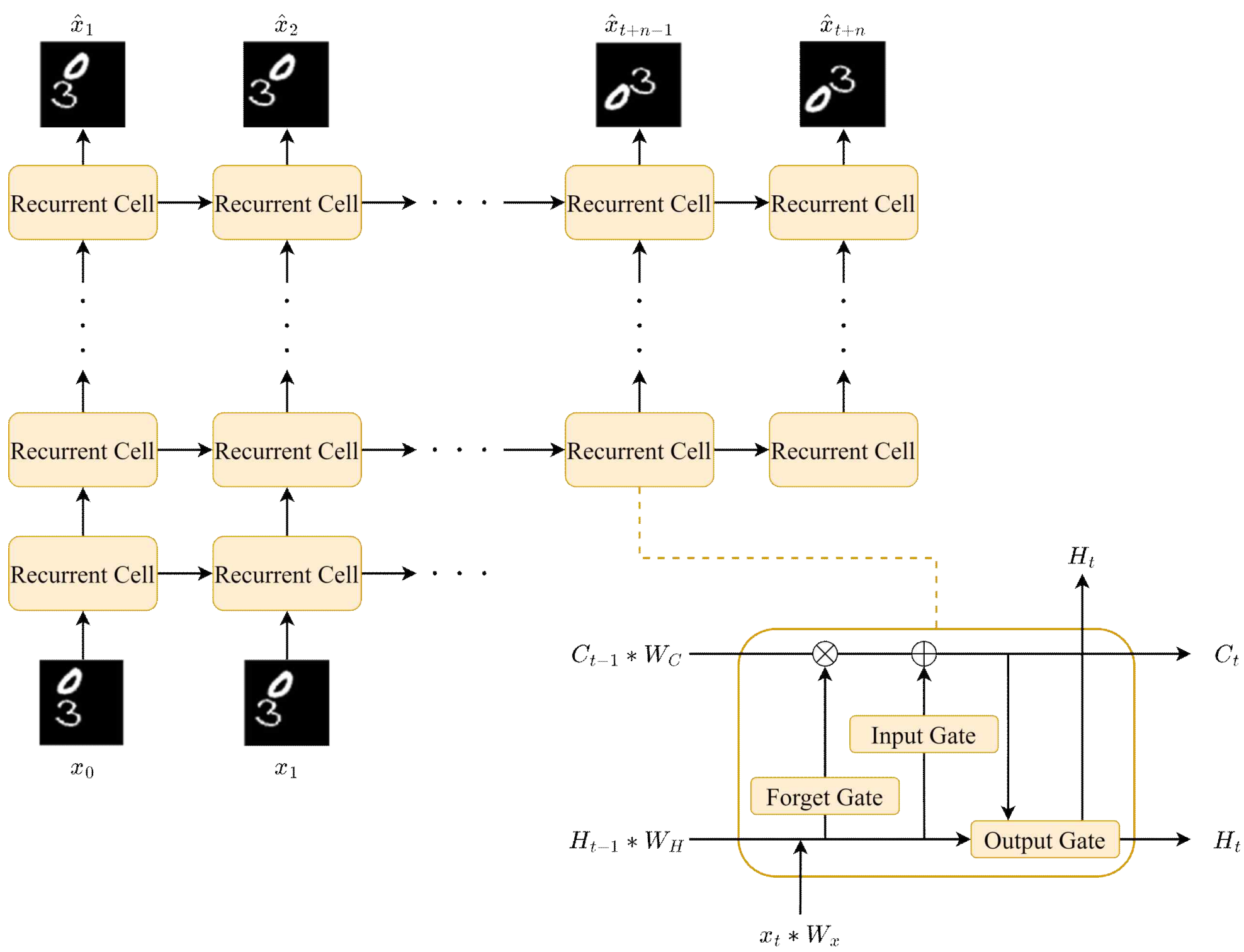

Fig. 2.1. General structure of Recurrent-based SISO models.

Initially, Shi et al. [15] showed interesting results by proposing ConvLSTM. ConvLSTM extended the concept of FC-LSTM [16] and effectively performed prediction by combining CNN and LSTM. It showed much lower parameters and excellent performance compared to the method combining fully connected with LSTM. This results is due to the properties that CNN is powerful in the computer vision field compared to fully connected. After this, Recurrent-based SISO model structure based on ConvLSTM has become the mainstream in the spatiotemporal prediction field.

Wang et al. proposed PredRNN [17], PredRNN++ [18], Eldetic 3D LSTM (E3D-LSTM) [19], and Memory In Memory (MIM) [20]. PredRNN utilizes the output of last recurrent cell as the input of first recurrent cell in the next time step. This vertically updates the state $C$ and $H$, which were previously updated horizontally along the time step and improves performance. PredRNN++ adds separate memory to the recurrent cell to update the state in both horizontal and vertical. In addition, the authors inspired by the work of Shrivastava et al. [38], integrates Gradient Highway Unit (GHU) into the $H$, output of the output gate in the first recurrent cell to avoid long-term gradient loss that occurs in deep networks and further improve prediction performance. E3D-LSTM integrates 3D CNN into RNNs instead of 2D CNN to update high-dimensional states and improve long-term prediction performance. MIM uses a cascading module to exploit the differences between adjacent cyclic states and effectively capture non-stationary information.

On the other hands, recent studies integrates interesting methods. Lin et al. [21] integrates self-attention into ConvLSTM and proposed SA-LSTM. Integrated self-attention memories allows to focus on important features and capture global spatiotemporal information. Yu et al. [22] proposed CrevNet, a conditionally reversible network based on CNN. Guen and Thome [23] proposed PhyDNet, an insteresting model that improves performance by utilizing physical knowledge. Lee, et al [24] performs long-term predictions while minimizing bottlenecks through alignment learning and memory query decomposition. Chang et al. [12] proposed  Motion Aware Unit (MAU), composed of attention and fusion modules to capture motion information. Tang et al. [25] proposed novel recurrent cell named SwinLSTM by integrating Swin Transformer [39] and LSTM to overcome the limitations of CNN that difficulty captures global information. Recently, a novel recurrent cell, named VMRNN [26] that integrates Vision Mamba [40] was proposed.

These models effectively capture spatiotemporal information using interesting methods and achieve excellent results. However, RNNs used to capture temporal information have a fatal flaw in that they are inherently limited in parallelism and take a long time to learn, making them inefficient.

B. Recurrent-free MIMO models

Recurrent-free MIMO models use a structure consisting of encoder-decoder as shown in Fig. 2.2, and generally uses CNN and excludes RNNs. Spatial encoder and decoder reduce and restore size, repectively with extract spatial features of $x_{\text{in}}$. Temporal module learns spatiotemporal information based on reduced data by spatial encoder. In the case of ViT-based MIMO models, spatial encoder and decoder perform patch embedding and patch restoration. Temporal module learns important spatiotemporal information through ViT blocks. This method takes data corresponding to the entire time step at once, and returns the prediction data for a specific time step. It was proposed recently and has attracted attention due to its efficiency and competitive performance.

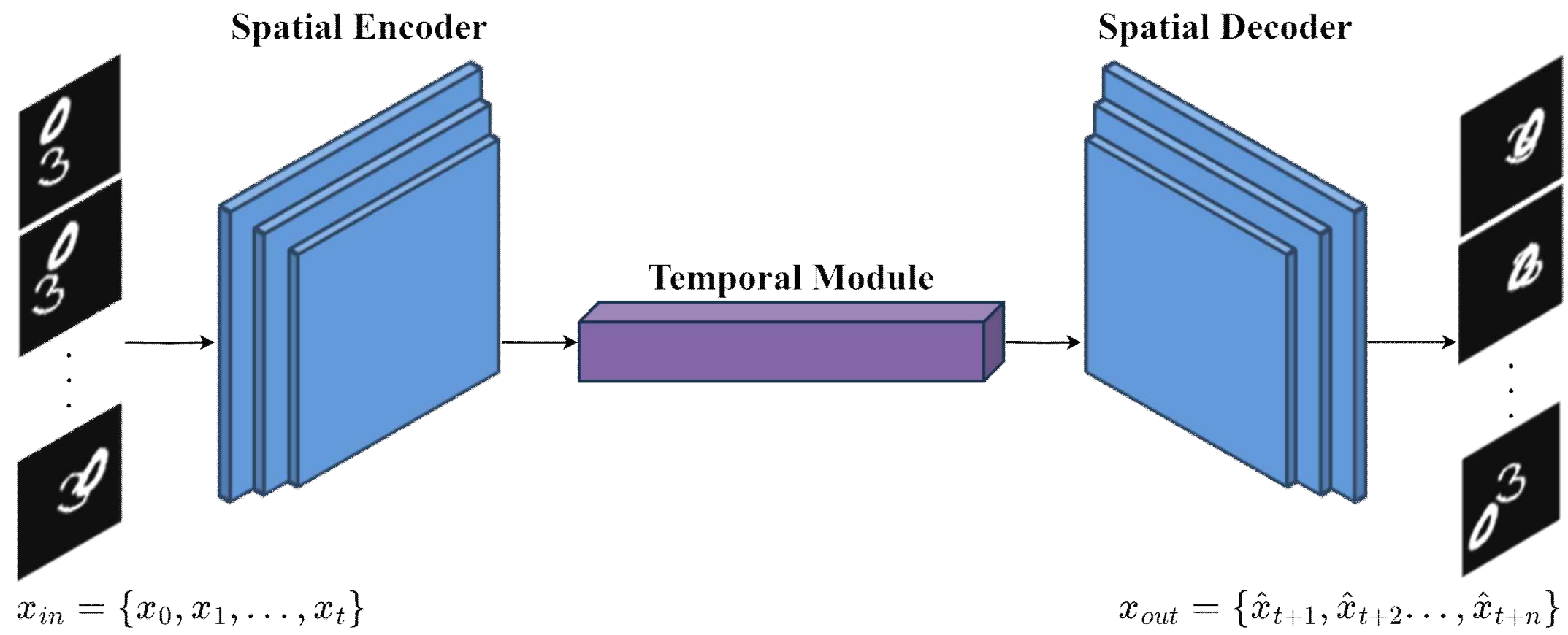

Fig. 2.2. General structure of Recurrent-free MIMO models.

1) CNN-based

Liu et al. [11] proposed TrajectoryCNN, that can capture and predict action dynamics in trajectory space. Gao et al. [32] proposed SimVP. SimVP reshape the data and placed a translator that can capture spatiotemporal features between encoder and decoder to provide fast and improved predict performance with only 2D CNN. Due to its remarkable efficiency and performance, proposed encoder-decoder architecture has become mainstream in the field of recurrent-free MIMO models. Tan et al. [28] improve SimVP by proposing Temporal Attention Unit (TAU). TAU captures static and dynamic information through attention blocks and improves long-term prediction performance. Zhong et al. [41] proposed Motion-Matrix based Video Prediction (MMVP) to infer motion while preserving the consistency of objects contained in spatiotemporal data. And recently, Nie et al. [42] proposed Wavelet-based Spatiotemporal (WaST) framework, which leveraging wavelet transform to extrach both low and high-frequency component in spatiotemporal data. WaST achieves state-of-the-art performance by utilizing the wavelet-domain high-frequency focal loss to preventing the disadvantage of mean squared error. On the other hand, Zhu et al. [43] proposed Faster yet Accurate Video Prediction (FAVP), designed for resource-constrained environment. FAVP replaces large-kernel convolution by utilizing dynamic kernel, Involution [44], and achieves efficiency. In addition, Tan et al. [45] improve SimVP by proposing SimVPv2. SimVPv2 goes beyond SimVP by replacing the translator consisting of inception U-Net with gated Spatiotemporal Attention (gSTA) module, which achieves remarkable performance improvements inspired by success of large-size kernel CNNs.

2) ViT-based

Ye et al. [30] proposed a ViT-based spatiotemporal prediction approach initially. It achieves competitive results and demonstrates the potential of ViT-based spatiotemporal prediction. Ning et al. [31] proposed MIMO-VP and emphasizes the strength of recurrent-free MIMO models by using transformer architecture consisting of local space-time blocks and multiple output decoders. And recently, Nie et al. [1] proposed Triplet Attention Module (TAM) to improve both accuracy and efficiency. TAM also achieves remarkable results through three attention blocks, which emphasize important features and suppress unnecessary features in temporal, spatial, and channel, respectively.

Although these recurrent-free MIMO models are attracting much attention due to their accuracy and high efficiency, there is a still room for improvement as they have been proposed relatively recently. In particular, CNNs, while fast, difficulty to capture global information, whereas ViTs, despite their accuracy, have difficulty capturing local information and limitations from high-computational complexity due to the self-attention.

### C. Advanced CNN

For preventing the limitations from CNNs, many variant structures have been proposed. Dai et al. [46] proposed Deformable Convolutional Networks (DCN) for extract features from not only fixed regions but also flexible regions. DCN apply offsets per each pixel for flexible sampling. These learnable offsets indicate dynamic sampling location. Wang et al. [47] proposed Kervolution to replace linear operations of convolution with nonlinear kernel functions, which provides richer expressive power. Li et al. [44] proposed Involution, which models spatial heterogeneity by applying dynamic kernels at different locations. It demonstrates usefulness through comparative experiments on various fields, including classification, detection, and segmentation tasks.

There have been recent attempts to overcome local limitations by changing the structure rather than transforming the inner operation. ConvNeXt [48, 49] modernizes ResNet [50] with ViT's design approach and shows performance beyond ViTs. Guo et al. [51] proposed Visual Attention Network (VAN). It overcome the local properties of CNNs with large-size kernel and decomposite it for efficiency. Furthermore, Hou et al. [52] combine transformer-style CNNs with large-size kenel based on simplifed self-attention mechanism. Proposed Conv2Former shows effective global receptive field and demonstrates the usefulness of advanced CNNs.

Inspired by these advances in the image-related computer vision fields, our goal is to improve the recurrent-free MIMO model by applying and optimizing an accurate yet efficient Transformer-style CNNs for spatiotemporal prediction. Specifically, in this paper, we design a patch embedding which consists of multiple convolution layers, and two attention blocks to capture both local and global information without mixing the time and channel axes.

# 3. METHOD

## A. Preliminaries

Given spatiotemporal data $x_{\text{in}} = \{x_{0,} ..., x_t\}$ with multiple time steps $t$, the goal is to predict the next $n$ time steps $x_{\text{out}} = \{x_{t+1}, ..., x_{t+n}\}$. Each input and output data is represented as a 5-dimensional tensor: $x_{\text{in}} \in \mathbb{R}^{\text{B} \times \text{T} \times \text{C} \times \text{H} \times \text{W}}$ and $x_{\text{out}} \in \mathbb{R}^{\text{B} \times \text{T} \times \text{C} \times \text{H} \times \text{W}}$. $B$, $T$, $C$, $H$, and $W$ denote batch size, number of time steps, number of channels, height, and width, respectively. Model $F$ learns to predict $\hat{x}$ by minimizing the discrepancy between the prediction $\hat{x}$ and $x_{\text{out}}$ using the loss function $\ell$, as formulated in (1):

$$\begin{aligned} \hat{x} &= F(x_{\text{in}}) \\ w^* &= \underset{w}{\operatorname{argmin}}\, \ell(\hat{x}, x_{out}) \end{aligned} \tag{1}$$

Where $w$ denotes the weights of the model $F$, which are optimized by minimizing the loss function $\ell$.

## B. MIMO-ESP structure

We propose a novel accurate yet efficient recurrent-free MIMO model, MIMO-ESP. The overall structure of MIMO-ESP is shown in Fig. 3.1. Where Ⓡ denotes reshape process. It consists of patch embed and back, and $N$ blocks. To prevent weakness of ViT, which difficulty to capture local information, we replace simple composition of patch embed and patch back to multiple convolution layers. Furthermore, to applying and optimizing advanced CNNs for spatiotemporal prediction.

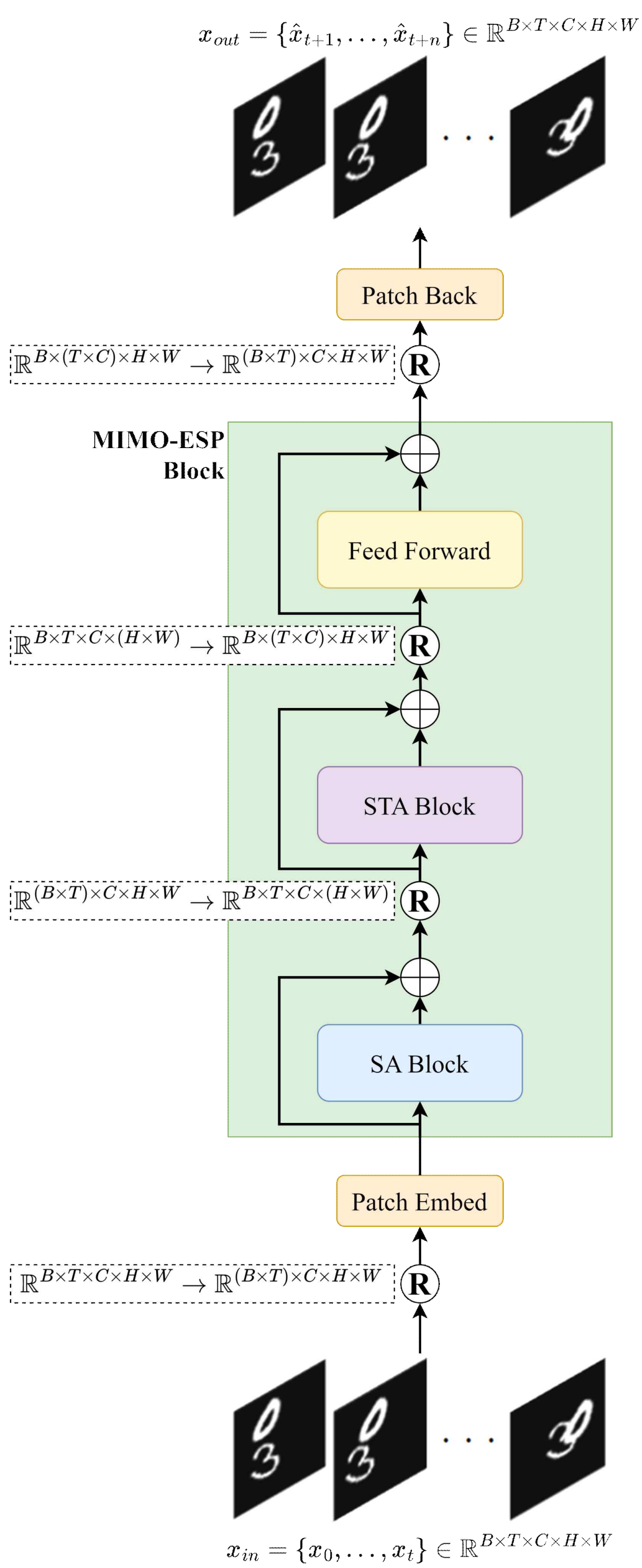

Fig. 3.1. Overall structure of MIMO-ESP.

Specifically, two attention blocks, Spatial Attention Block (SA Block) and Spatiotemporal Attention Block (STA Block) to capture spatial and spatiotemporal information was applied. For efficient spatiotemporal prediction without 3D convolution layers, each attention block performed after reshape process and it designed for preserving the time and channel axes independently. Each components of MIMO-ESP is carefully designed to enhance both local and global information capture, thereby improving spatiotemporal prediction performance. The following subsections present a detailed description of their design and functionality.

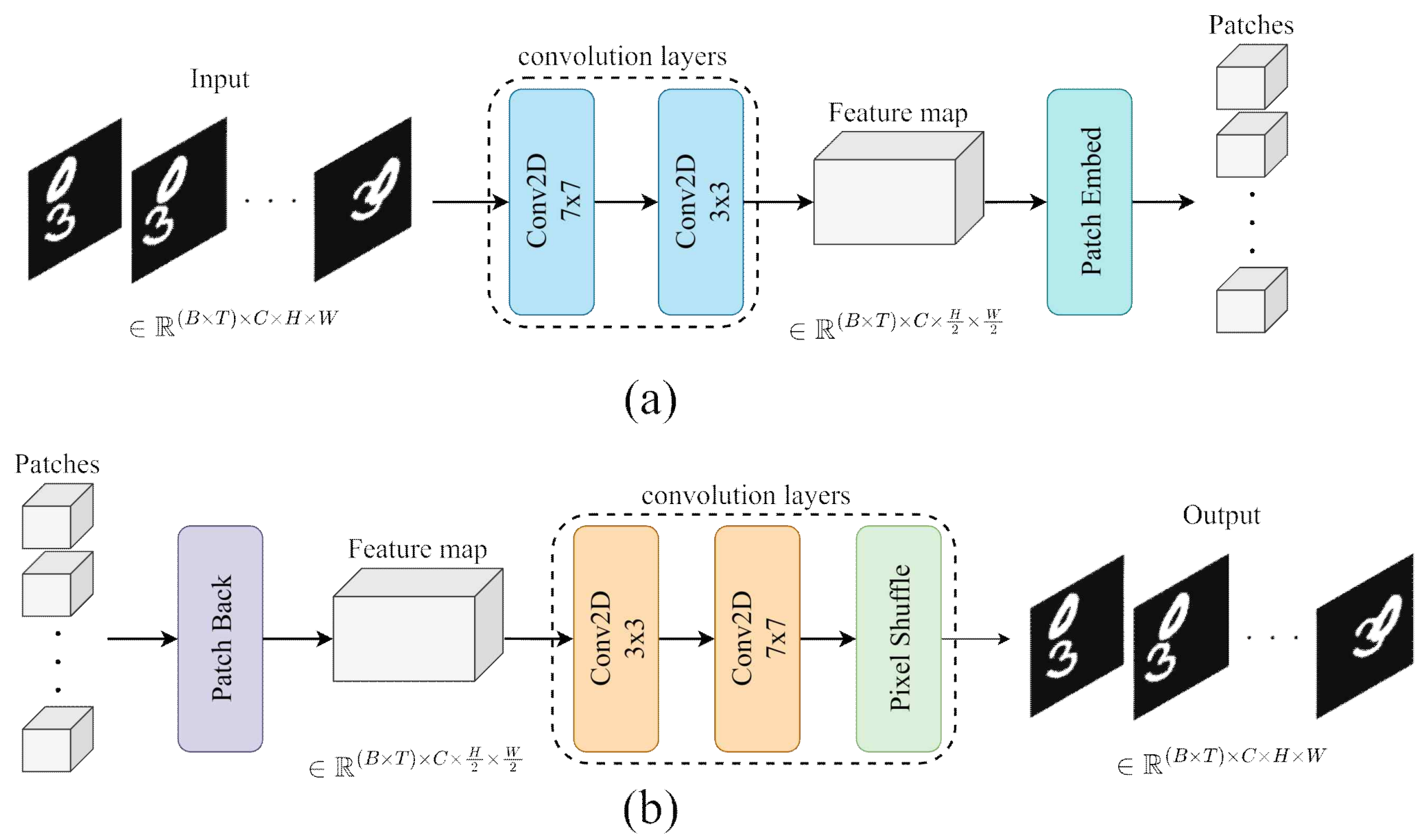

Fig. 3.2. Patch embed and back structure.
((a): patch embed process, (b): patch back process.)

1) Patch embed and back with multiple convolution layers

Yuan et al. [53] mentioned the vulnerability of ViTs, which difficulty in capturing local spatial information and occurs hardness at initial process of training in middle size dataset like ImageNet. To prevent this, patch embed and back process, after reshape process in MIMO-ESP consist of three convolution layers as shown in Fig. 3.2. Reshape process change the shape of the 5-dimensional tensor $x_{in} \in \mathbb{R}^{B\times T\times C\times H\times W}$ to 4-dimensional tensor $x_{in} \in \mathbb{R}^{(B\times T)\times C\times H\times W}$ by multiplying time axis $T$ with batch size $B$. Where convolution layers in Fig. 3.2 (a) extract local information and reduce the size of the input data. On the other hands, convolution layers in Fig. 3.2 (b) restore the size with local information and make the output data. Convolution layers are employed with 7×7 size kernel with stride 2, and 3×3 size kernel, respectively. This ensure in extracting local information over a wide range of receptive field. After this, patch embed was performed efficienctly and the dimension of the input data is expanded. In addition, this process reduces the size of the input data, thereby reducing the load on spatiotemporal prediction tasks with two attention blocks.

Patch back is performed in the reverse order of the patch embed process. In addition, pixel shuffle approach was applied for restoring the size of data. Similary, extended dimension is reduced through two stages, patch back and convolution layers, either and patches restored through patch back.

2) Spatial attention block

MIMO-ESP is based on two attention blocks, SA Block and STA Block, which extracts the core information. The structure of SA Block is shown in Fig. 3.3. Where Norm, PWConv, DWConv, and ⊙ denotes layer normalization, point-wise convolution, depth-wise convolution, and hadamard product, respectively. To perform spatiotemporal prediction efficiently, SA Block treats spatiotemporal data similarly with image, following the output shape from patch embed process and apply 2D convolution layers. SA Block apply Transformer-style CNNs for capture global information. This architecture effectively increases the receptive fields and allows the model to focus on important feature. If given the input data $x$, the process of SA Block can be represented as (2):

$$\begin{aligned} x_{att} &= LayerNorm(x) \\ A &= PWConv(x_{att}) \\ V &= DWConv(\sigma(PWConv(x_{att})) \\ SABlock(x) &= PWConv(A \odot V) \end{aligned} \quad (2)$$

Where $\sigma$ denotes the GELU activation function. DWConv is employed with 7×7 size kernel. The output of the SA Block process $x_{SA}$ including patch embed with reshape is represented as (3):

$$\begin{aligned} x_S &= Reshape(PatchEmbed(x_{in})) \\ x_{SA} &= SABlock(x_S) \oplus x_S \end{aligned} \quad (3)$$

Where $\oplus$ denotes the element-wise add. In SA Block, DWConv operates independently with channels of the input data $x_S$. Therefore, with large kernel CNNs, it can capture the global spatial correlations of each patch, embedded by the patch embed process. On the other hand, PWConv operates independently with height and width, only consider the channel information. This effectively complements the values of each patch.

3) Spatiotemporal attention block

The structure of STA Block is shown in Fig. 3.4. Where DW-DConv denotes depthwise-dilated convolution. To preserve time information without mixing channel information, reshape process applied before the STA Block. Specifically, 4-dimensional tensor, which output of the SA Blo ck $x_{SA} \in \mathbb{R}^{(B \times T) \times C \times H \times W}$ reshaped to $x_{ST} \in \mathbb{R}^{B \times T \times C \times (H \times W)}$. Similary with flattened patches in ViTs, height and width in $x_{SA}$ flattened for representing spatiotemporal data with 4-dimensional tensor and apply 2D convolution layers. As a result, spatiotemporal information including time $T$, height $H$, and width $W$ are formatted like image, without mixing channel information $C$. In addition, to efficiently capture the time information, DW-DConv consists of multiple convolution layers with dilation applied only for time axis. The dilation applied to each convolution layer increase gradually to efficiently capture long-term time information. The detailed structure of DW-D Conv is shown in Fig. 3.4. The process of STA Block can be represented as (4):

$$
\begin{aligned}
&x_{att} = LayerNorm(x) \\
&A = PWConv(x_{att}) \\
&V_{ST} = DW-DConv(\sigma(PWConv(x_{att})) \\
&STABlock(x) = PWConv(A \odot V_{ST})
\end{aligned}
\tag{4}
$$

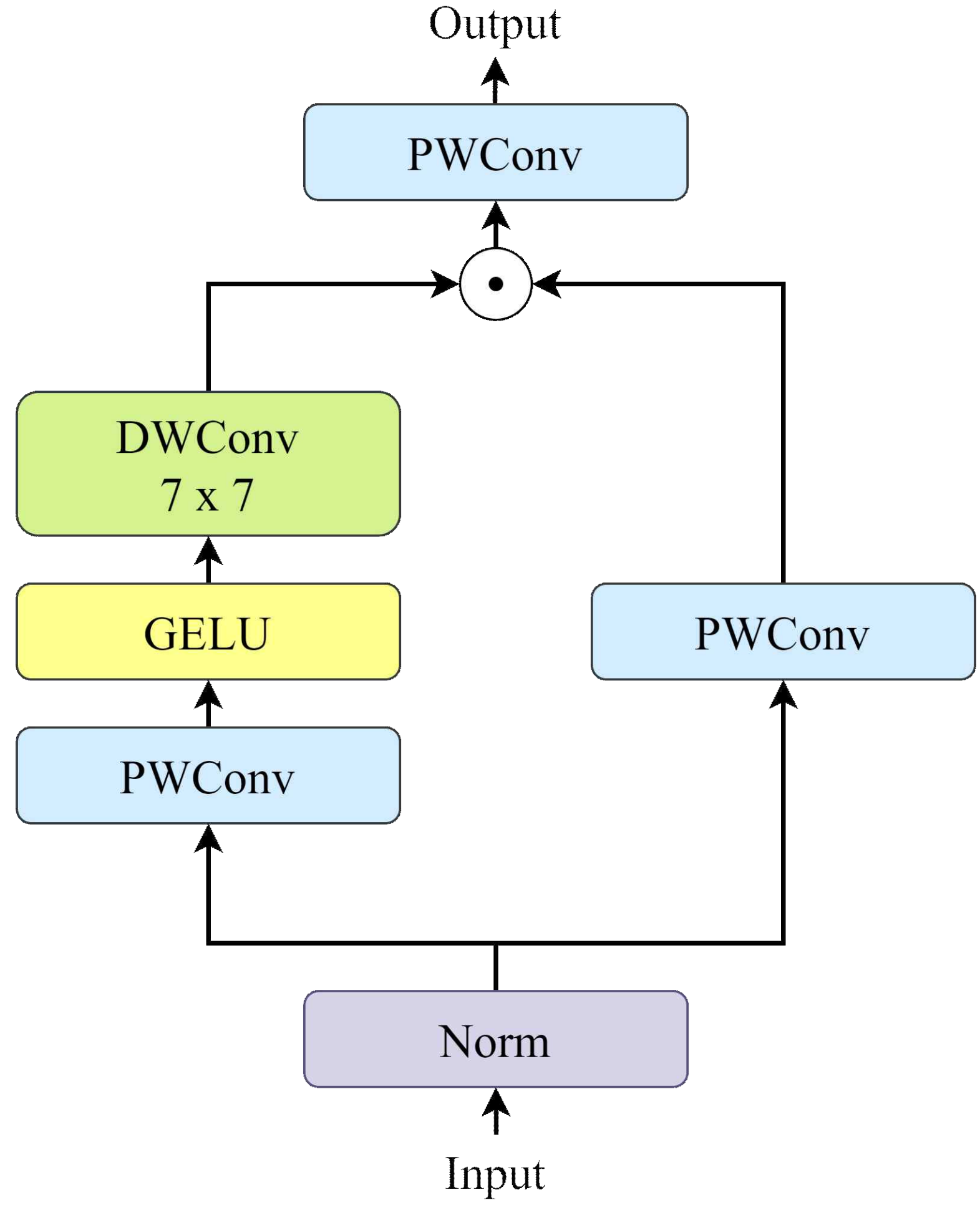

Fig. 3.3. SA Block structure.

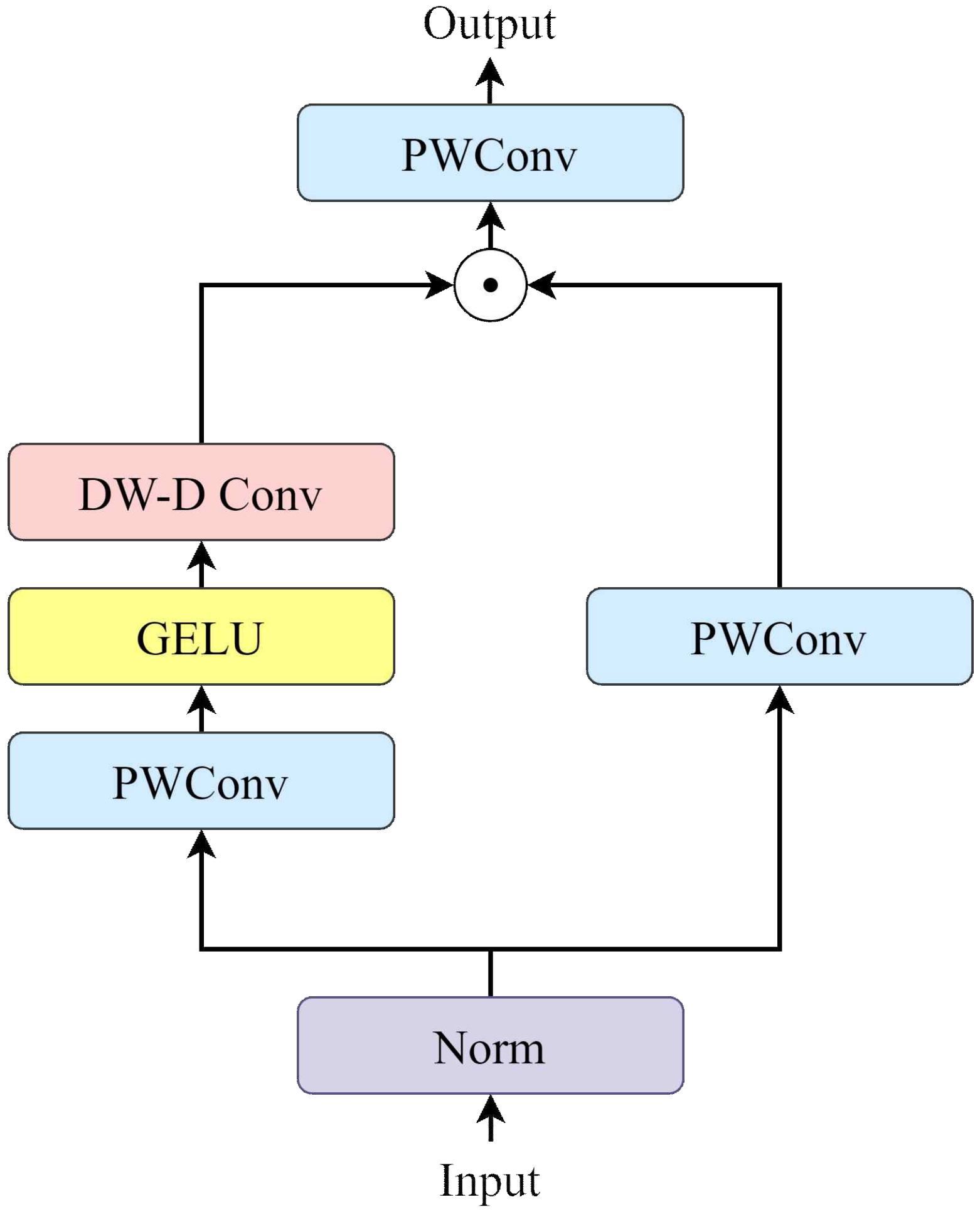

Fig. 3.4. STA Block structure.

Different with SA Block process, instead of DWConv, there is DW-D Conv which contains multiple convolution layers with increasing dilation.

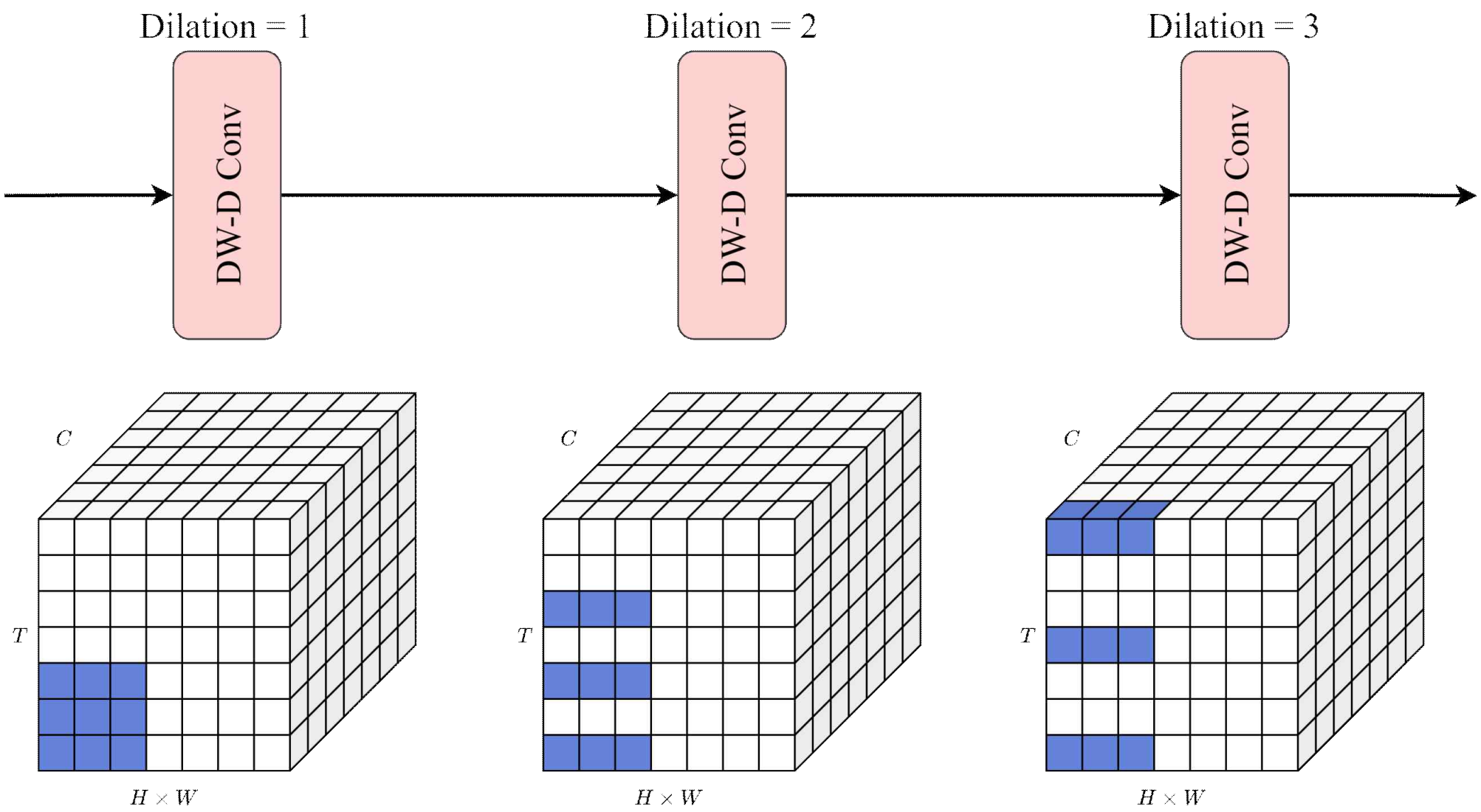

Fig. 3.5. Example of increased receptive field by DW-D Conv.

The output of the STA Block process $x_{STA}$ including reshape process is represented as (5):

$$x_{ST} = Reshape(x_{SA})$$
$$x_{STA} = STABlock(x_{ST}) \oplus x_{ST} \quad (5)$$

In addition to the spatial information extracted by the SA Block, reshaped format and DW-D Conv of the STA Block effectively extracts spatiotemporal information, enabling more accurate spatiotemporal prediction. Local information captured by the patch embed composed of multiple convolution layers, global information is captured by two attention blocks, SA Block and STA Block, which effectively improves the respective limitations of CNN and ViT-based recurrent-free MIMO models.

Finally, predicted value $x_{out}$, the output of MIMO-ESP $F(x_{in})$ represented as follows (6):

$$
\begin{aligned}
&x_{FF} = Reshape(x_{STA}) \\
&MLP = PWConv(\sigma(PWConv(x_{FF}))) \\
&FeedForward(x_{FF}) = MLP \oplus x_{FF} \\
&F(x_{in}) = PatchBack(Reshape(FeedForward(x_{FF})))
\end{aligned}
\tag{6}
$$

Feed Forward consisting of two PWConv and plays a role similar to the Multi-Layer Perceptron (MLP). It effectively complements the information derived through two attention blocks with the reshape process, increases and restores the channel, and stabilizes the output of the MIMO-ESP. In addition, reshape process re-applied before patch back, which makes $FeedForward(x_{FF}) \in \mathbb{R}^{B \times (T \times C) \times H \times W}$ to the tensor, which can be represented in 4-dimensional space, $\in \mathbb{R}^{(B \times T) \times C \times H \times W}$. Patch back complements the local spatiotemporal information through multiple convolution layers and patch shuffle and restores the patches to spatiotemporal data.

# 4. EXPERIMENTAL SETTINGS

A. Dataset descriptions

To comprehensively evaluate the prediction performance of MIMO-ESP, we conduct extensive experiments using three promising spatiotemporal benchmark datasets. These datasets were selected to evaluate spatiotemporal prediction performance in various scenarios, including video, traffic flow, and precipitation prediction:

- MovingMNIST [16] is a synthetic video dataset, which consists of two digits. Each digit moving independently on the grid map which consists of resolution 64×64×1. This dataset performed representative benchmark role in the field of spatiotemporal prediction. Following the previous work [23, 32], we use generated 10,000 data samples for the train and original 10,000 data samples for test. Each frame normalized between 0 and 1, and designed to get the 10 input frames and predict 10 frames. Train was performed during 2,000 epochs.
- TaxiBJ [7] is a real-world dataset collected from taxi in the Beijing area. Each time step consists of a grid map with the size 32×32×2, representing the traffic volumes and each channel represents inflow and outflow of the traffic. Following the previous work [20, 23]. we normalize it between 0 and 1, and design the experiment to get the input previous 4 time step and predict 4 future time steps during 100 epochs. We use 20,461 and 500 data samples for train and test, respectively.
- Radar-Echo[1] is a public dataset that consists of radar echo images in the form of temporal sequences and is widely used in precipitation

1 https://doi.org/10.5281/zenodo.7059116

prediction tasks. Each radar echo image has a resolution of 100×100×1. We converted the image sequences into a spatiotemporal format using a sliding window. Specifically, the train data consists of five time steps, while the test data consists of ten time steps to evaluate the performance for flexible-term prediction scenarios. A total of 2,700 samples are split into a 7:3 ratio, with 1,890 samples used for train and 810 for test. Train was performed during 50 epochs.

We summarize and describe the configurations for each dataset in Table 1. Where $T_{in}$ and $T_{out}$ denote the number of time steps for input and predict data, respectively. Furthermore, $N_{train}$ and $N_{test}$ denote the number of samples for train and test, respectively.

Table 1. Configurations for each dataset.

| Dataset | Resolution | $T_{in}$ | $T_{out}$ | $N_{train}$ | $N_{test}$ | Epochs |
|---|---|---|---|---|---|---|
| MovingMNIST | (64,64,1) | 10 | 10 | 10000 | 10000 | 2000 |
| TaxiBJ | (32,32,2) | 4 | 4 | 20461 | 500 | 100 |
| Radar-Echo | (100,100,1) | 5 | 10 | 1890 | 810 | 50 |

B. Metrics

We evaluate the quality of prediction results using three different metrics, including Mean Squared Error (MSE), Mean Absolute Error (MAE), and Structural Similarity Index Measure (SSIM). Error metrics, MSE and MAE are calculated pixel-wise, with lower values indicate better performance. In addition, to evaluate similarity, SSIM is used, with higher values indicate better performance. We report pixel-wise MSE, MAE, and SSIM for two experiments, which use MovingMNIST and TaxiBJ dataset. Each metrics can be represented as (7):

$$MSE = \frac{1}{T}\sum_{t=1}^{T}(y_t - \hat{y}_t)^2 \tag{7}$$
$$MAE = \frac{1}{T}\sum_{t=1}^{T}\left|y_t - \hat{y}_t\right|$$
$$SSIM = \frac{1}{T}\sum_{t=1}^{T} l(y_t, \hat{y}_t) * c(y_t, \hat{y}_t) * s(y_t, \hat{y}_t)$$

Where $y_t$ and $\hat{y}_t$ denote ground-truth and predicted value for the frame at time step $t$. In addition, $l()$, $c()$,and $s()$ in SSIM equation denote the functions for calculating luminance, contrast, and structure, respectively.

Furthermore, we set metrics by referring to related prior study [54] to evaluate the performance of precipitation prediction. We additionally report Heidke Skill Score (HSS) [55] and Critical Success Index (CSI) metrics. Below are the four terms used to compute these metrics which defined in Chen et al. [56]:

- True Positive (TP): The sample labels that are correctly identified by the model as positive.
- True Negative (TN): The sample labels that are correctly identified by the model as negative.
- False Positive (FP): The negative sample labels that are incorrectly identified by the model as positive.
- False Negative (FN): The positive sample labels that are incorrectly identified by the model as negative.

Finally, HSS and CSI are defined in (8):

$$HSS = \frac{2(TP \times TN - FN \times FP)}{(TP + FN)(FN + TN) + (TP + FP)(FP + TN)} \tag{8}$$
$$CSI = \frac{TP}{TP + FN + FP}$$

Specifically, ground truth and prediction values are converted to 0 and 1, by the specific threshold value. We calculated with three threshold values, 5, 20, and 40.  After the calculate, we report average HSS and CSI values.

### C. Implementation details

Experiments were conducted on Python 3.10.8 and Pytorch 2.1.1 with Intel(R) Core(TM) i7-10700 CPU @ 2.90GHz and RTX 3070 in Windows 10 WSL environments. Batch size is set to 16, AdamW optimizer and L1+L2 loss function are used to consider both MAE and MSE. Learning rate of 0.0001, 0.001, and 0.001 are applied to video prediction with MovingMNIST dataset, traffic prediction with TaxiBJ dataset, and precipitation prediction with Radar-Echo dataset, respectively.

The hyperparameters of the comparison models follow those set in their paper. For MIMO-ESP, hyperparameter settings per each dataset are as summarized in Table 2.

Table 2. Hyperparameter settings for each dataset.

| Dataset | Batch size | Learning rate | $N_{block}$ | Patch size | Embed hid | Embed dim | Dilations |
|---|---|---|---|---|---|---|---|
| MovingMNIST | 16 | 1e-4 | 8 | 2 | 64 | 128 | (1,2,4) |
| TaxiBJ | 16 | 1e-3 | 4 | 4 | 32 | 64 | (1,2) |
| Radar-Echo | 16 | 1e-3 | 4 | 4 | 128 | 128 | (1,2) |

Where $N_{block}$, embed hid, embed dim, and dilations are denote number of the MIMO-ESP Block,  hidden dimensions in convolution layers for patch embed and back, output dimensions in patch embed and back, and applied dilation for STA Block, respectively. The hyperparameters are carefully designed considering the amount and resolution of each dataset, as well as the number of time steps.

# 5. EXPERIMENTAL RESULTS

In this section, we report both quantatitive and qualitative experimental results and provide a detailed analysis of them. In quantitative comparison results, bold in each table indicates best performance and underline represents second. To evaluate the accuracy yet efficiency of the proposed MIMO-ESP, we conduct comprehensive comparative experiments, including the results from various SISO and MIMO models.

### A. Result on MovingMNIST dataset

First, we compare the synthetic video prediction performance on MovingMNIST dataset, which perform the representative benchmark role. Where the model takes ten input frames and predicts next ten frames. As shown in Table 1, our MIMO-ESP achieved the best performance compared to other recent models. As mentioned in experimental setting section, lowest MSE and MAE indicate that the prediction result with MIMO-ESP most close with groud truth and it also means there's no significant difference. SSIM value in Table 1 also indicate that the prediction result with MIMO-ESP has highest quality and most similar with ground truth. In particular, compared to recent models, MSE, MAE, and SSIM improved by approximately 10%. The qualitative results consistently demonstrate the effectiveness of MIMO-ESP, as shown in Fig. 4.1. Where G.T denotes ground truth and error calculated with $|G.T-Preds|$. In error map, brighter regions indicate higher discrepancies from the ground truth. Compared with TAU, MIMO-ESP can accurately tracks the movement of objects in spatiotemporal data by capturing both local and global information.

Table 3. Quantitative comparision results on MovingMNIST dataset.

| Method | MSE↓ | MAE↓ | SSIM↑ |
|---|---|---|---|
| ConvLSTM [15] | 103.3 | 182.9 | 0.707 |
| PredRNN [17] | 56.8 | 126.1 | 0.867 |
| PredRNN++ [18] | 46.5 | 106.8 | 0.898 |
| E3D-LSTM [19] | 41.3 | 87.2 | 0.910 |
| MIM [20] | 44.2 | 101.1 | 0.910 |
| SA-ConvLSTM [21] | 43.9 | 94.7 | 0.913 |
| CrevNet [22] | 22.3 | - | 0.949 |
| PhyDNet [23] | 24.4 | 70.3 | 0.947 |
| LMC-Memory [24] | 41.5 | - | 0.924 |
| SimVP [32] | 23.8 | 68.9 | 0.948 |
| VPTR-NAR [30] | 63.6 | - | 0.882 |
| MIMO-VP [31] | <u>17.7</u> | <u>51.6</u> | <u>0.964</u> |
| SwinLSTM [25] | <u>17.7</u> | - | 0.962 |
| MMVP [41] | 22.2 | - | 0.952 |
| TAU [28] | 19.8 | 60.3 | 0.957 |
| **MIMO-ESP (Ours)** | **16.4** | **49.9** | **0.966** |

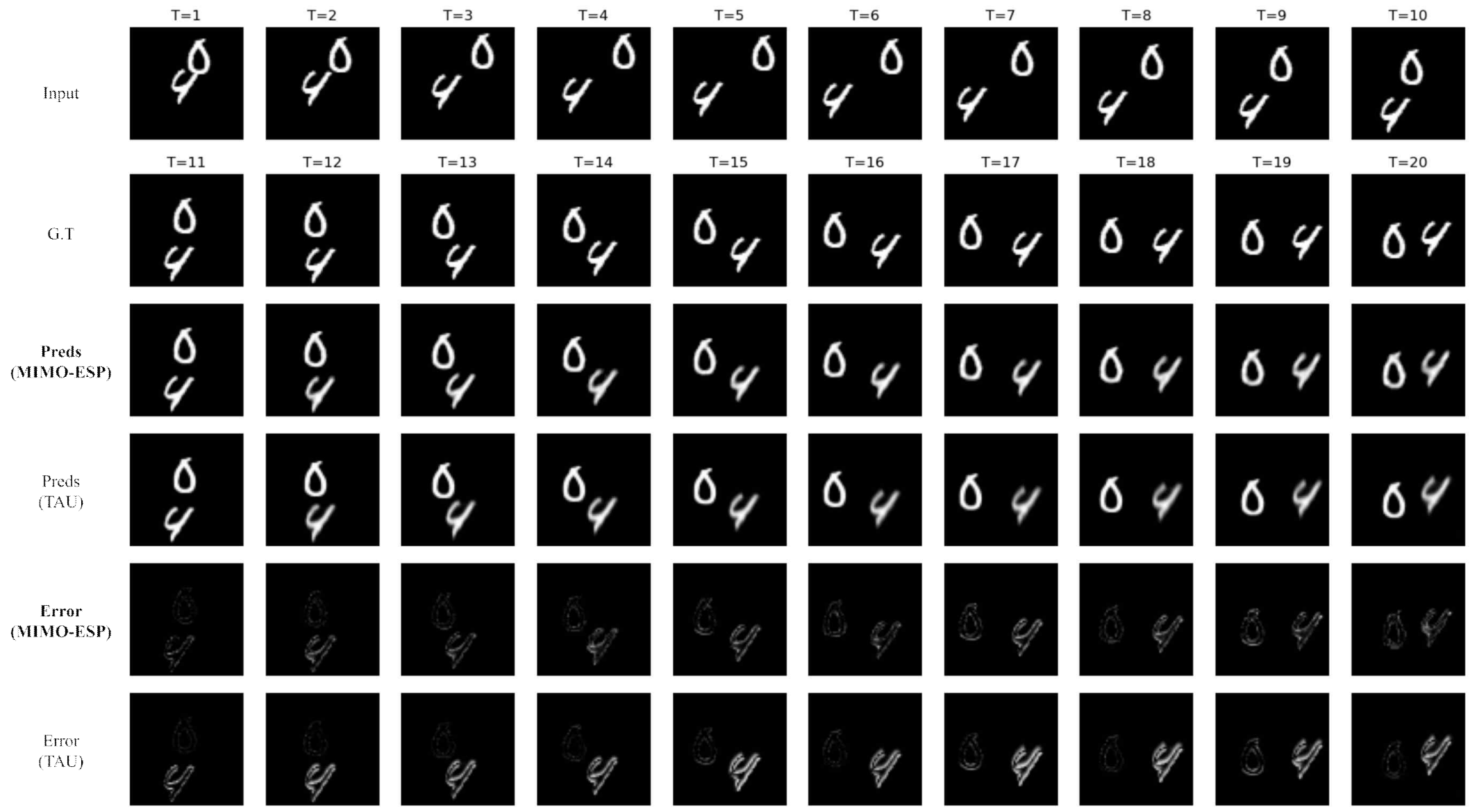

Fig. 5.1. Qualitative comparison results on MovingMNIST dataset.

B. Result on TaxiBJ dataset

In traffic prediction, we compare the short-term prediction performance, which inputs four time steps and predicts the next four time steps. Traffic flow prediction is challenging due to the importance of specific locations, such as highways or buildings, and spatiotemporal heterogeneity. Nevertheless, as shown in Table 2, MIMO-ESP achieves competitive results. Especially when compared to other recurrent-based SISO and recurrent-free MIMO models, MIMO-ESP shows remarkable performance and is competitive considering its efficient structure compared to recurrent-based SISO models. This is consistent with the experiments on the MovingMNIST dataset, which again emphasizes the effectiveness of MIMO-ESP.

Table 4. Quantitative comparision results on TaxiBJ dataset.

| Method | MSE↓ | MAE↓ | SSIM↑ |
|---|---|---|---|
| ConvLSTM [15] | 48.5 | 17.7 | 0.978 |
| PredRNN [17] | 46.4 | 17.1 | 0.971 |
| PredRNN++ [18] | 44.8 | 16.9 | 0.977 |
| E3D-LSTM [19] | 43.2 | 16.9 | 0.979 |
| MIM [20] | 42.9 | 16.6 | 0.971 |
| SA-ConvLSTM [21] | 38.9 | - | - |
| PhyDNet [23] | 41.9 | 16.2 | 0.982 |
| SimVP [32] | 41.4 | 16.2 | 0.982 |
| SwinLSTM [25] | 39.0 | - | 0.980 |
| TAU [28] | <u>34.4</u> | <u>15.6</u> | <u>0.983</u> |
| MIMO-ESP (Ours) | 32.6 | 15.3 | 0.984 |

The qualitative comparison results are shown in Fig. 4.2. Similarly with Fig. 4.1, G.T denotes ground truth and error calculated with $|G.T-Preds|$ and brighter region in error map indicate higher discrepancies from the ground truth. Despite the difficulty of traffic prediction, MIMO-ESP showed remarkable ability. Compared to TAU, where the brightness of the error map is high in a specific region, MIMO-ESP captures global information also, and error values equivalent evenly.

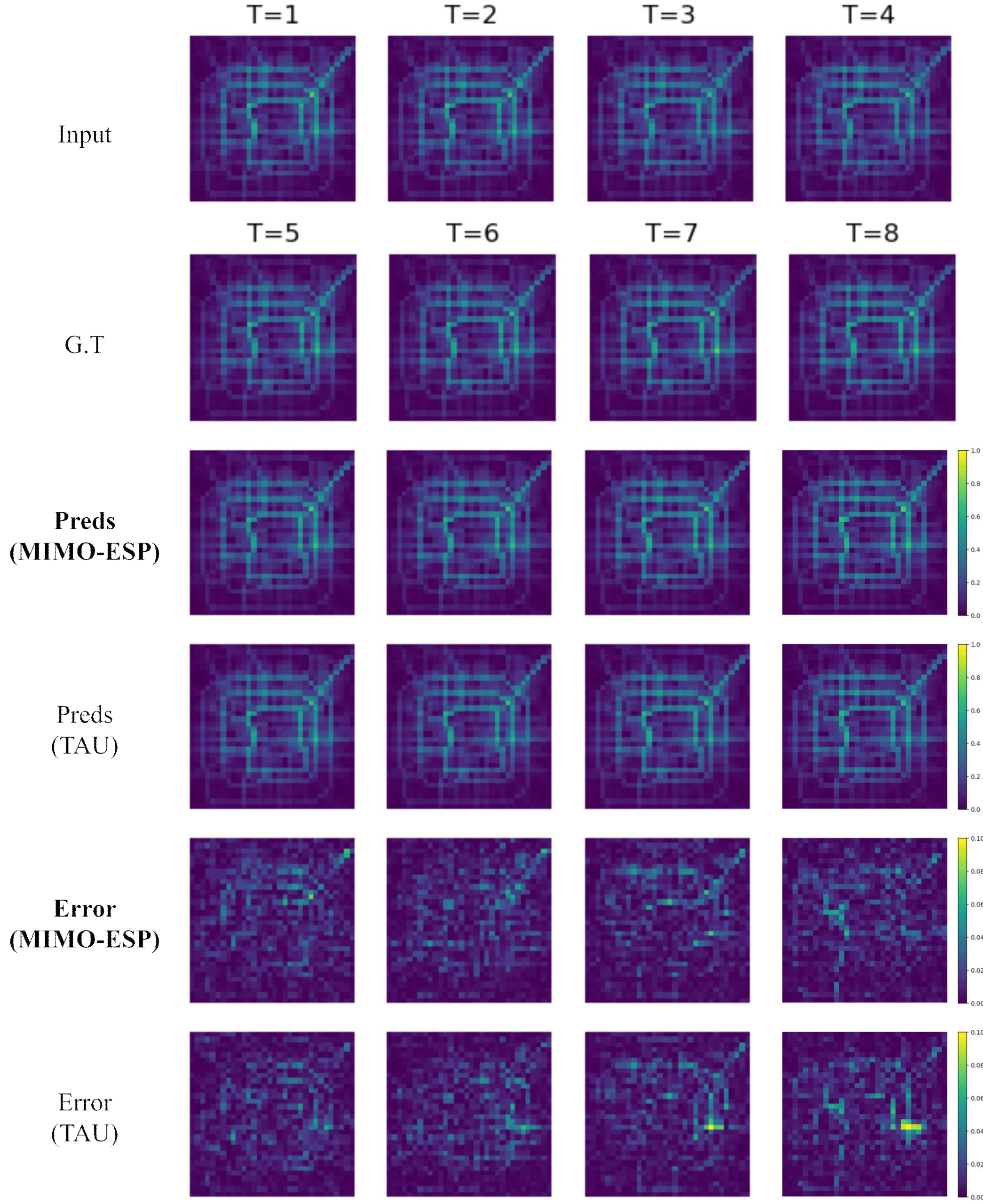

Fig. 5.2. Qualitative comparison results on TaxiBJ dataset.

C. Result on Radar-Echo dataset

In precipitation prediction, we also consider the flexible-term prediction performance, which inputs five time steps and predicts ten time steps. To perform flexible prediction, MIMO models use an autoregressive approach. This approach can be seen as part of a Multi-In-Single-Out (MISO) structure, which predicts the same time steps as the input data and uses it as input again to perform flexible prediction.

We report quantitative comparison results on Table 3. Our MIMO-ESP achieves competitive results consistently. The results for MSE, MAE, and SSIM emphasize the utility of MIMO-ESP, as well as its robustness as a flexible prediction approach with different input and prediction time steps. In particular, the results for HSS and CSI show that our MIMO-ESP not only performs structurally similar predictions, but also has higher accuracy for precipitation predictions than other models, making it suitable for a wide range of applications.

Table 5. Quantitative comparision results on Radar-Echo dataset.

| Method | HSS↑ | CSI↑ | MSE↓ | MAE↓ | SSIM↑ |
|---|---|---|---|---|---|
| ConvLSTM [15] | 0.697 | 0.673 | <u>9.06</u> | 190.74 | <u>0.830</u> |
| PredRNN [17] | 0.699 | 0.681 | 9.55 | <u>182.36</u> | 0.823 |
| PredRNN++ [18] | 0.678 | 0.653 | 10.09 | 199.05 | 0.819 |
| MIM [20] | 0.694 | 0.671 | 9.12 | 190.80 | <u>0.830</u> |
| PhyDNet [23] | 0.685 | 0.658 | 10.51 | 211.52 | 0.802 |
| SimVP [32] | 0.700 | 0.671 | 9.39 | 200.23 | 0.818 |
| MIMO-VP [31] | <u>0.704</u> | <u>0.682</u> | 9.77 | 191.29 | 0.829 |
| SwinLSTM [25] | 0.646 | 0.602 | 13.49 | 258.45 | 0.721 |
| TAU [28] | 0.691 | 0.669 | 9.92 | 202.07 | 0.819 |
| MIMO-ESP (Ours) | 0.705 | 0.683 | 8.97 | 180.02 | 0.837 |

The qualitative comparison results are shown in Fig 4.3. Compared to the previous two experimental results on MovingMNIST and TaxiBJ, we observed that the quality of predictions after the fifth time step significantly decreased due to the accumulated error. This result demonstrate the effectiveness of the MIMO structure emphasized in this paper and previous studies. Despite these challenges, MIMO-ESP continues to perform competitive predictions. Due to the accumulated errors, MIMO-ESP also fails to maintain detailed representations after the fifth time step. However, it captures the differences in brightness across regions by considering global information. This demonstrates the remarkable effectiveness of the MIMO-ESP approach when compared to TAU, which gradually blurs the prediction and struggles to capture the differences in brightness.

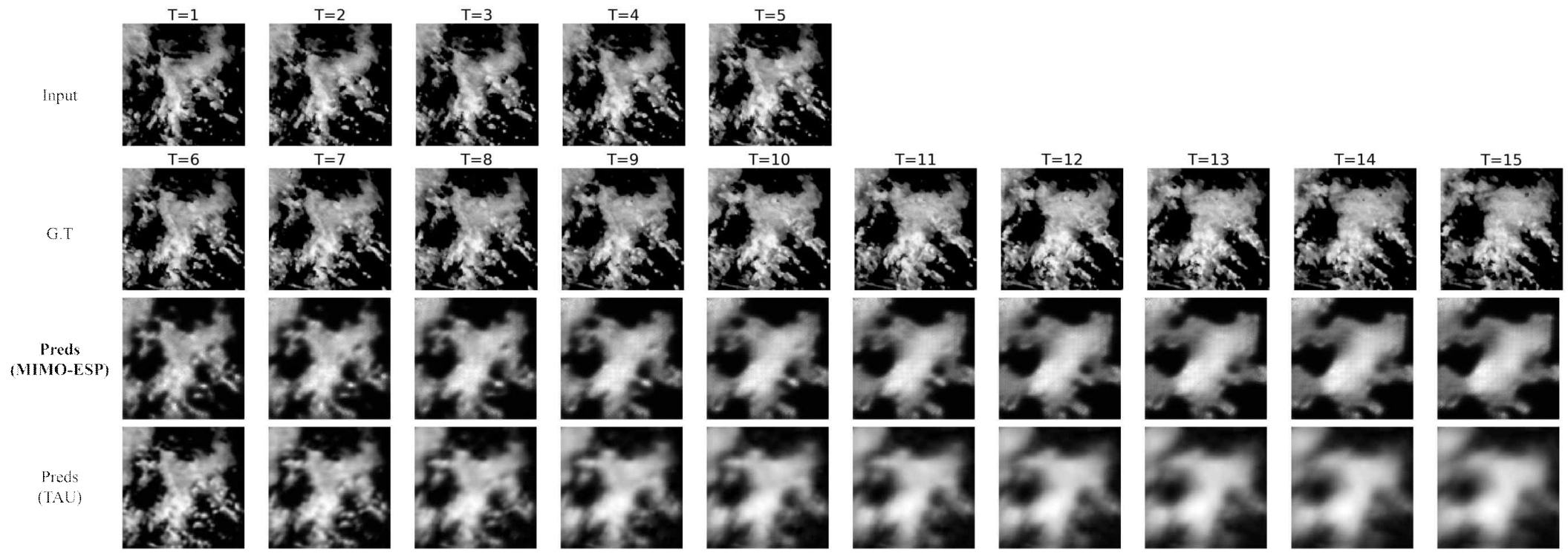

Fig. 5.3. Qualitative comparison results on Radar-Echo dataset.

D. Efficiency comparison

Accurate yet efficient spatiotemporal prediction models are essential for real-world applications. To evaluate the efficiency of MIMO-ESP, we compare its computational cost and model size with four representative recurrent-free MIMO models including two CNN-based models [32, 28] and two ViT-based models [30, 31]. Specifically, we report two things, number of parameters (Params) and floating point operations (FLOPs). In the table, smaller Params indicate a smaller model size, indicating that spatiotemporal features can be learned effectively even with a small size.

On the other hand, high FLOPs indicate a high computational load of the model, indicating that the structure is complex and requires a lot of computing resources for prediction.

As shown in Table 4, MIMO-ESP achieves competitive efficiency when compared with ViT-based recurrent-free MIMO models. Although model size that represented with Params approximately two times more than one of the promising model MIMO-VP on experiment on MovingMNIST, FLOPs which represent complexity and required computational resource approximately reduced 98%. In addition, when compared with VPTR-NAR, both Params and FLOPs are grately reduced. Considering MIMO-ESP's accurate prediction performance, its accuracy yet efficiency was emphasized.

Table 6. Efficiency comparison with ViT-based models.

| Method | MovingMNIST | | Radar-Echo | |
|---|---|---|---|---|
| | Params(M) | FLOPs(G) | Params(M) | FLOPs(G) |
| VPTR-NAR [30] | 165.8 | 197.00 | - | - |
| MIMO-VP [31] | **22.6** | 961.00 | 1.93 | 6.19 |
| **MIMO-ESP (Ours)** | 48.7 | **16.61** | **0.77** | **1.74** |

Quantitative efficiency comparison with CNN-based recurrent-free MIMO models are shown in Table 4. MIMO-ESP also achieves competitive performance in here. When checked efficiency on MovingMNIST dataset, there is no significant difference in both Params and FLOPs. In addition, when checked on TaxiBJ and Radar-Echo dataset, MIMO-ESP achieves most efficient performance. Considering its accurate performance based-on capturing not only local but also global information, it emphasizing the effectiveness and promising of our approach.

Table 7. Efficiency comparison with CNN-based models.

| Method | MovingMNIST | | TaxiBJ | | Radar-Echo | |
|---|---|---|---|---|---|---|
| | Params(M) | FLOPs(G) | Params(M) | FLOPs(G) | Params(M) | FLOPs(G) |
| SimVP [32] | 58.0 | 19.43 | 13.79 | 13.61 | 13.59 | 8.88 |
| TAU [28] | **44.7** | **15.96** | 9.55 | 2.49 | 4.10 | 2.89 |
| MIMO-ESP (Ours) | 48.7 | 16.61 | **1.99** | **0.28** | **0.77** | **1.74** |

E. Ablation study

To further analyze the effectiveness of each component of MIMO-ESP, we conduct two types of ablation studies by changing each components and evaluating its impact with MovingMNIST dataset. First, we replace the SA Block and STA Block of MIMO-ESP with the self-attention block used in ViT-based MIMO models [30, 31]. We report FLOPs to evaluate the efficiency of the proposed Transformer-style CNN in the spatiotemporal prediction tasks. Specifically, when w/o. denotes without and replaced with self-attention block.

Table 8. Ablation study on efficiency with MovingMNIST dataset.

| | Params(M) | FLOPs(G) |
|---|---|---|
| w/o. SA and STA Block | **16.8** | 45.63 |
| w/o. SA Block | 18.1 | 26.80 |
| w/o. STA Block | 18.6 | 40.05 |
| MIMO-ESP (Ours) | 48.7 | **16.61** |

As shown in Table 6, when each block, SA and STA Block, is replaced with self-attention block, the Params decreases but the complexity increases significantly. This is consistent with the efficiency comparison with the previous ViT-based MIMO models, and emphasizes the efficiency of the proposed approach. In particular, this feature is further emphasized when the STA Block, which processes temporal information in addition to spatial, is replaces.

Next, we report the changes in MSE and MAE, and evaluate the impact of the proposed approach on the performance. Table 7 shows the performance depending on the presence or absense of dilation in STA Block, preserving time information without mixing channel information, and multiple convolution layers in patch embed and back, respectively. Where w/o. denotes without and replace to base CNN structure.

Table 9. Ablation study on performance with MovingMNIST dataset.

| | MSE↓ | MAE↓ |
|---|---|---|
| w/o. dilation in STA Block | 37.8 | 106.26 |
| w/o. preserving time information | 35.3 | 88.37 |
| w/o. convolution layers in patchfy | 23.9 | 67.13 |
| MIMO-ESP (Ours) | 16.4 | 49.93 |

While the removal of dilation in STA Block, degrades prediction performance, confirming importance of dilation in efficient capturing long-term time information. When we apply reshape process similar with previous recurrent-free MIMO models, which can not preserving time information, observing significant variations in both MSE and MAE. It emphasizing the effectiveness of our novel reshape approaches. Finally, when removal multiple convolution layers in patchfy, observing minor variations and also emphasizing the importance of not only global but also local information in spatiotemporal prediction.

## F. Discussion

In this paper, we proposed MIMO-ESP, a novel accurate yet efficient recurrent-free MIMO model. Our MIMO-ESP can consider both local and global information based-on transformer-style CNN with multiple convolution layers in patchfy process. In addition, novel reshape process which can preserving time information without mixing channel, and STA Block which applied dilation contribute to significantly improves the performance.

Despite remarkable accuracy yet efficiency of MIMO-ESP, its application in real-world problems, warrants careful analysis. As shown in the efficiency comparison, although MIMO-ESP achieves competitive efficiency and novel performance when considering its accurate prediction, as the amount of dataset increases both Params and FLOPs also increased significantly. This suggests that our MIMO-ESP can make accurate prediction based on its novel components but potentially limiting in when required huge amount of dataset environments. To clear this, our future research directions focus on optimizing the structure including comprehensive evaluate with various spatiotemporal datasets.

# 6. CONCLUSION

We proposed novel accurate yet efficient recurrent-free MIMO model, MIMO-ESP. To improving MIMO models, we designed the novel structure that optimized for spatiotemporal prediction including novel patchfy and reshape process, two attention blocks including SA Block and STA Block. Specifically, novel patchfy and reshape process allows MIMO-ESP to capture local information well and more accurate prediction based on prevent mixing time and channel information. Two attention blocks based on transformer-style CNN, SA Block and STA Block allows MIMO-ESP to capture both global spatial and spatiotemporal information. In addition, dilation which applied to the STA Block enables efficient expansion and takes into account long-term spatiotemporal information.

We demonstrates the usefulness of MIMO-ESP and potential of proposed approach through comparison experiments using three dataset, MovingMNIST, TaxiBJ, and Radar-Echo. Our MIMO-ESP achieves both novel accuracy and competitive efficiency. Although there is room for warrants careful analysis when application in real-world problems, the results of ablation study demonstrate and emphasize the potential and effectiveness for each component of MIMO-ESP. We expect that this study will further stimulate interest in improvement recurrent-free MIMO models and provide new insights for designing.

# References

[1] Nie, Xuesong, et al. "Triplet attention transformer for spatiotemporal predictive learning." Proceedings of the IEEE/CVF Winter Conference on Applications of Computer Vision. 2024.

[2] Oprea, Sergiu, et al. "A review on deep learning techniques for video prediction." IEEE Transactions on Pattern Analysis and Machine Intelligence 44.6 (2020): 2806-2826.

[3] Amato, Federico, et al. "A novel framework for spatio-temporal prediction of environmental data using deep learning." Scientific reports 10.1 (2020): 22243.

[4] Bai, Cong, et al. "Rainformer: Features extraction balanced network for radar-based precipitation nowcasting." IEEE Geoscience and Remote Sensing Letters 19 (2022): 1-5.

[5] Gao, Zhihan, et al. "Earthformer: Exploring space-time transformers for earth system forecasting." Advances in Neural Information Processing Systems 35 (2022): 25390-25403.

[6] Wu, Hao, et al. "Earthfarsser: Versatile spatio-temporal dynamical systems modeling in one model." Proceedings of the AAAI Conference on Artificial Intelligence. Vol. 38. No. 14. 2024.

[7] Zhang, Junbo, Yu Zheng, and Dekang Qi. "Deep spatio-temporal residual networks for citywide crowd flows prediction." Proceedings of the AAAI conference on artificial intelligence. Vol. 31. No. 1. 2017.

[8] Saleh, Khaled, Artur Grigorev, and Adriana-Simona Mihaita. "Traffic accident risk forecasting using contextual vision transformers." 2022 IEEE 25th International Conference on Intelligent Transportation Systems (ITSC). IEEE, 2022.

[9] Grigorev, Artur, Khaled Saleh, and Adriana-Simona Mihaita. "Traffic Accident Risk Forecasting using Contextual Vision

Transformers with Static Map Generation and Coarse-Fine-Coarse Transformers." 2023 IEEE 26th International Conference on Intelligent Transportation Systems (ITSC). IEEE, 2023.

[10] Kong, Yu, Zhiqiang Tao, and Yun Fu. "Deep sequential context networks for action prediction." Proceedings of the IEEE conference on computer vision and pattern recognition. 2017.

[11] Liu, Xiaoli, et al. "Trajectorycnn: a new spatio-temporal feature learning network for human motion prediction." IEEE Transactions on Circuits and Systems for Video Technology 31.6 (2020): 2133-2146.

[12] Chang, Zheng, et al. "Mau: A motion-aware unit for video prediction and beyond." Advances in Neural Information Processing Systems 34 (2021): 26950-26962.

[13] Finn, Chelsea, Ian Goodfellow, and Sergey Levine. "Unsupervised learning for physical interaction through video prediction." Advances in neural information processing systems 29 (2016).

[14] Sharma, Vijeta, et al. "Video processing using deep learning techniques: A systematic literature review." IEEE Access 9 (2021): 139489-139507.

[15] Shi, Xingjian, et al. "Convolutional LSTM network: A machine learning approach for precipitation nowcasting." Advances in neural information processing systems 28 (2015).

[16] Srivastava, Nitish, Elman Mansimov, and Ruslan Salakhudinov. "Unsupervised learning of video representations using lstms." International conference on machine learning. PMLR, 2015.

[17] Wang, Yunbo, et al. "Predrnn: Recurrent neural networks for predictive learning using spatiotemporal lstms." Advances in neural information processing systems 30 (2017).

[18] Wang, Yunbo, et al. "Predrnn++: Towards a resolution of the deep-in-time dilemma in spatiotemporal predictive learning." International conference on machine learning. PMLR, 2018.

[19] Wang, Yunbo, et al. "Eidetic 3D LSTM: A model for video prediction and beyond." International conference on learning representations. 2018.

[20] Wang, Yunbo, et al. "Memory in memory: A predictive neural network for learning higher-order non-stationarity from spatiotemporal dynamics." Proceedings of the IEEE/CVF conference on computer vision and pattern recognition. 2019.

[21] Lin, Zhihui, et al. "Self-attention convlstm for spatiotemporal prediction." Proceedings of the AAAI conference on artificial intelligence. Vol. 34. No. 07. 2020.

[22] Yu, Wei, et al. "Efficient and information-preserving future frame prediction and beyond." International Conference on Learning Representations. 2020.

[23] Guen, Vincent Le, and Nicolas Thome. "Disentangling physical dynamics from unknown factors for unsupervised video prediction." Proceedings of the IEEE/CVF conference on computer vision and pattern recognition. 2020.

[24] Lee, Sangmin, et al. "Video prediction recalling long-term motion context via memory alignment learning." Proceedings of the IEEE/CVF Conference on Computer Vision and Pattern Recognition. 2021.

[25] Tang, Song, et al. "Swinlstm: Improving spatiotemporal prediction accuracy using swin transformer and lstm." Proceedings of the IEEE/CVF International Conference on Computer Vision. 2023.

[26] Tang, Yujin, et al. "Vmrnn: Integrating vision mamba and lstm for efficient and accurate spatiotemporal forecasting." Proceedings of the IEEE/CVF Conference on Computer Vision and Pattern Recognition. 2024.

[27] Bai, Shaojie, J. Zico Kolter, and Vladlen Koltun. "An empirical evaluation of generic convolutional and recurrent networks for

sequence modeling." arXiv preprint arXiv:1803.01271 (2018).

[28] Tan, Cheng, et al. "Temporal attention unit: Towards efficient spatiotemporal predictive learning." Proceedings of the IEEE/CVF Conference on Computer Vision and Pattern Recognition. 2023.

[29] Audibert, Julien, et al. "Usad: Unsupervised anomaly detection on multivariate time series." Proceedings of the 26th ACM SIGKDD international conference on knowledge discovery & data mining. 2020.

[30] Ye, Xi, and Guillaume-Alexandre Bilodeau. "Vptr: Efficient transformers for video prediction." 2022 26th International Conference on Pattern Recognition (ICPR). IEEE, 2022.

[31] Ning, Shuliang, et al. "MIMO is all you need: A strong multi-in-multi-out baseline for video prediction." Proceedings of the AAAI conference on artificial intelligence. Vol. 37. No. 2. 2023.

[32] Gao, Zhangyang, et al. "Simvp: Simpler yet better video prediction." Proceedings of the IEEE/CVF conference on computer vision and pattern recognition. 2022.

[33] Nie, Xuesong, et al. "Wavelet-Driven Spatiotemporal Predictive Learning: Bridging Frequency and Time Variations." Proceedings of the AAAI Conference on Artificial Intelligence. Vol. 38. No. 5. 2024.

[34] Dosovitskiy, Alexey. "An image is worth 16x16 words: Transformers for image recognition at scale." arXiv preprint arXiv:2010.11929 (2020).

[35] Luo, Wenjie, et al. "Understanding the effective receptive field in deep convolutional neural networks." Advances in neural information processing systems 29 (2016).

[36] Lee, Jinyoung, and Gyeyoung Kim. "CNN-Based Time Series Decomposition Model for Video Prediction." IEEE Access (2024).

[37] Vaswani, A. "Attention is all you need." Advances in Neural Information Processing Systems (2017).

[38] Srivastava, Rupesh K., Klaus Greff, and Jürgen Schmidhuber. "Training very deep networks." Advances in neural information processing systems 28 (2015).

[39] Liu, Ze, et al. "Swin transformer: Hierarchical vision transformer using shifted windows." Proceedings of the IEEE/CVF international conference on computer vision. 2021.

[40] Zhu, Lianghui, et al. "Vision mamba: Efficient visual representation learning with bidirectional state space model." arXiv preprint arXiv:2401.09417 (2024).

[41] Zhong, Yiqi, et al. "Mmvp: Motion-matrix-based video prediction." Proceedings of the IEEE/CVF international conference on computer vision. 2023.

[42] Nie, Xuesong, et al. "Wavelet-driven spatiotemporal predictive learning: bridging frequency and time variations." Proceedings of the AAAI Conference on Artificial Intelligence. Vol. 38. No. 5. 2024.

[43] Zhu, Junhong, et al. "Towards faster yet accurate video prediction for resource-constrained platforms." Neurocomputing 611 (2025): 128663.

[44] Li, Duo, et al. "Involution: Inverting the inherence of convolution for visual recognition." Proceedings of the IEEE/CVF conference on computer vision and pattern recognition. 2021.

[45] Tan, Cheng, et al. "SimVPv2: Towards simple yet powerful spatiotemporal predictive learning." IEEE Transactions on Multimedia (2025).

[46] Dai, Jifeng, et al. "Deformable convolutional networks." Proceedings of the IEEE international conference on computer vision. 2017.

[47] Wang, Chen, et al. "Kervolutional neural networks." Proceedings of

the IEEE/CVF conference on computer vision and pattern recognition. 2019.

[48] Liu, Zhuang, et al. "A convnet for the 2020s." Proceedings of the IEEE/CVF conference on computer vision and pattern recognition. 2022.

[49] Woo, Sanghyun, et al. "Convnext v2: Co-designing and scaling convnets with masked autoencoders." Proceedings of the IEEE/CVF conference on computer vision and pattern recognition. 2023.

[50] He, Kaiming, et al. "Deep residual learning for image recognition." Proceedings of the IEEE conference on computer vision and pattern recognition. 2016.

[51] Guo, Meng-Hao, et al. "Visual attention network." Computational visual media 9.4 (2023): 733-752.

[52] Hou, Qibin, et al. "Conv2former: A simple transformer-style convnet for visual recognition." IEEE transactions on pattern analysis and machine intelligence (2024).

[53] Yuan, Li, et al. "Tokens-to-token vit: Training vision transformers from scratch on imagenet." Proceedings of the IEEE/CVF international conference on computer vision. 2021.

[54] Luo, Chuyao, et al. "A novel LSTM model with interaction dual attention for radar echo extrapolation." Remote Sensing 13.2 (2021): 164.

[55] Hogan, Robin J., et al. "Equitability revisited: Why the “equitable threat score” is not equitable." Weather and Forecasting 25.2 (2010): 710-726.

[56] Chen, Haoyu, and Kyungbaek Kim. "Multi-convolutional channel residual spatial attention u-net for industrial and medical image segmentation." IEEE Access (2024).

# 효율적인 시공간 예측을 위한 합성곱 신경망 기반 다중 입·출력 모델

진 현 석

전남대학교대학원 인공지능융합학과

(지도교수 : 김경백)

(국문초록)

최근, 시공간 예측 분야에서 순환 신경망 기반 모델들의 한계를 극복하기 위해 합성곱 신경망 또는 트랜스포머 기반 모델들이 제안되었다. 이러한 모델들은 순차적 특성으로 인해 병렬화가 제한되는 비효율성과, 재귀적 예측 방법으로 인한 오류 누적 등 순환 신경망 기반 모델들의 본질적인 한계를 극복하고 높은 성능을 보였다. 그러나 여전히 몇 가지 한계가 존재한다. 첫 번째로, 합성곱 신경망 기반 모델들은 지역적인 정보만을 포착하는 커널의 한계로 인해 전역적인 정보를 고려하는 데 어려움이 있고, 성능이 제한적이다. 또한 시공간 데이터의 시간 차원과 채널 차원을 결합하여 처리하기 때문에, 두 차원의 정보가 섞인다는 한계가 존재한다. 두 번째로, 트랜스포머 기반 모델들은 셀프 어텐션 연산으로 인해 복잡도가 높고, 학습에 오랜 시간을 필요로 한다. 본 논문에서는 이러한 한계를 극복하기 위해 합성곱 신경망 기반 다중 입·출력 모델, MIMO-ESP라는 새로운 구조의 모델을 제안한다. 제안된 MIMO-ESP는 전역적인 정보를 고려하면서도, 합성곱 신경망을 기반으로 트랜스포머 아키텍처를 구성하여 셀프 어텐션 연산의 복잡도를 크게 개선할 수 있다. 또한 시간 차원을 결합하지 않고 독립적으로 취급 및 확장하여 시공간 정보를 효과적으로 고려할 수 있다. 이러한 구조는 제안된 MIMO-ESP를 효율적이면서도 높은 예측력을 유지할 수 있도록 한다. 비디

오, 교통 및 강수량 예측 작업을 포함하는 세 가지 유망한 벤치마크 데이터 세트를 사용하여 광범위한 실험을 수행한 결과, 제안된 MIMO-ESP는 기존 모델들의 예측 성능을 능가하는 동시에 경쟁력 있는 효율성을 달성하여 제안된 방법론의 유용성을 입증하였다. 나아가 절제 연구 결과는 MIMO-ESP 내 각 구성 요소들의 유용성을 입증하며, 제안된 접근 방식의 잠재력을 강조한다.